# Preserving Differential Privacy Between Features in Distributed Estimation

**Christina Heinze-Deml**[*]    heinzedeml@stat.math.ethz.ch
*Seminar for Statistics*
*ETH Zürich*
*8092 Zürich, Switzerland*

**Brian McWilliams**[*]    brian@disneyresearch.com
*Disney Research*
*8006 Zürich, Switzerland*

**Nicolai Meinshausen**    meinshausen@stat.math.ethz.ch
*Seminar for Statistics*
*ETH Zürich*
*8092 Zürich, Switzerland*

## Abstract

Privacy is crucial in many applications of machine learning. Legal, ethical and societal issues restrict the sharing of sensitive data making it difficult to learn from datasets that are partitioned between many parties. One important instance of such a distributed setting arises when information about each record in the dataset is held by different data owners (the design matrix is "vertically-partitioned").

In this setting few approaches exist for private data sharing for the purposes of statistical estimation and the classical setup of differential privacy with a "trusted curator" preparing the data does not apply. We work with the notion of $(\epsilon, \delta)$-distributed differential privacy which extends single-party differential privacy to the distributed, vertically-partitioned case. We propose PrIDE, a scalable framework for distributed estimation where each party communicates perturbed random projections of their locally held features ensuring $(\epsilon, \delta)$-distributed differential privacy is preserved. For $\ell_2$-penalized supervised learning problems PrIDE has bounded estimation error compared with the optimal estimates obtained without privacy constraints in the non-distributed setting. We confirm this empirically on real world and synthetic datasets.

## 1. Introduction

Data driven personalization—from user experience on the web to medicine and healthcare—relies on aggregating a large amount of potentially sensitive data relating to individuals from disparate sources in order to answer statistical queries. Understandably, from a privacy perspective it may be undesirable—or even impossible—for such data to be shared in an undisguised form. For example, in healthcare and medical science applications, highly personal information is collected about individuals which can be invaluable for diagnosis, treatment and drug discovery. The use and sharing of such data is governed by relevant laws such as the Health Insurance Portability and Accountability Act (HIPAA) which typically only allow data to be shared if it has been de-identified (Sarwate et al., 2014). However, even after a dataset has been sanitized, the risk of subjects being re-identified is an ongoing concern and in many such cases privacy breaches actually occurred (El Emam et al., 2011).

---

[*]. Both authors contributed equally to this work.
  A preliminary version of this work was presented at the NIPS 2016 workshop on private multi-party machine learning.

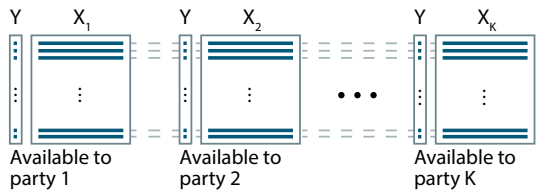

Figure 1: **Vertically partitioned data:** each party holds a subset of the total number of features, containing the data from the same set of individuals. Each party with access to $Y$ can estimate $\boldsymbol{\beta}_k$.

*Differential privacy* (DP) (Dwork, 2006) constitutes a powerful theoretical framework for guaranteeing that the output of a suitable algorithm will not allow the identification of individuals in a dataset. Recently, it has been considered as a method of complying with the many regulations for sharing data in e.g. healthcare applications (Dankar and El Emam, 2013). Informally, a differentially private algorithm is one that ensures information identifying an individual cannot be learned from the output of that algorithm on two datasets which differ only by that individual.[1]

In case of supervised learning, research has mainly focused on ensuring that a model estimated in the single party setting can be publicly released (Chaudhuri et al., 2011). However, in many application areas where sensitive data is held by several parties—e.g. health informatics, risk modelling and computational social science (D'Orazio et al., 2015)— estimating a model and performing statistical inference, rather than coefficient release, is often the stated goal. Therefore, an important open question concerns how sensitive data can be shared among different parties in a distributed computation framework to optimize a global statistical learning objective.

**Summary of contributions.** In §2 we formally introduce the problem setting—statistical estimation where *sensitive* data is partitioned *vertically* between multiple parties—and describe some of the unique challenges in this setting. In §3 we propose PRIDE (PRIvate Distributed Estimation), a scalable algorithm for differentially private statistical estimation when the data are partitioned vertically among multiple parties. Our key insight is that to ensure privacy, we require a small algorithmic change to the recently proposed DUAL-LOCO framework (Heinze et al., 2016). In §4, we show the following theoretical properties of PRIDE:

§4.1 **Privacy**: PRIDE preserves $(\epsilon, \delta)$-distributed differential privacy (cf. Definition 2).

---
1. Many definitions with subtle differences are used. We will formally state a definition for our purposes in §4.

§4.2 **Utility**: The estimation error of PRIDE with respect to the optimal coefficients (estimated in the non-distributed setting under no privacy constraints) is bounded.

The second main contribution is an extensive evaluation of the empirical behavior of PRIDE on a variety of simulated and real datasets in §5. We observe that PRIDE improves upon a fully-private baseline which avoids communicating any data between parties and quickly approaches the performance of the optimal solution. Related work is discussed in §6.

## 2. Problem setting

In this work, we are interested in objectives of the form

$$\min_{\mathbf{b} \in \mathbb{R}^p} \left\{ J(\mathbf{b}) := \frac{1}{n} \sum_{i=1}^n f_i(\mathbf{b}^\top \mathbf{x}_i) + \frac{\lambda}{2} \|\mathbf{b}\|_2^2 \right\} \tag{1}$$

where $\lambda > 0$ is the regularization parameter. The loss functions $f_i(\mathbf{b}^\top \mathbf{x}_i)$ depend on a response $y_i \in \mathbb{R}$ and linearly on the coefficients, $\mathbf{b} \in \mathbb{R}^p$ through a vector of covariates, $\mathbf{x}_i \in \mathbb{R}^p$. Furthermore, we assume all $f_i$ to be convex and smooth with Lipschitz continuous gradients. For example when $f_i(\mathbf{b}^\top \mathbf{x}_i) = (y_i - \mathbf{b}^\top \mathbf{x}_i)^2$, Eq. (1) corresponds to ridge regression; for logistic regression $f_i(\mathbf{b}^\top \mathbf{x}_i) = \log(1 + \exp(-y_i \mathbf{b}^\top \mathbf{x}_i))$. Let $\boldsymbol{\beta}$ denote the true underlying coefficients of interest.

In the multi-party setting where the data are vertically partitioned, each party $k$ has some proportion of the features corresponding to all of the observations (cf. Figure 1). Given a design matrix $\mathbf{X} \in \mathbb{R}^{n \times p}$ whose rows are $\mathbf{x}_i^\top$, each party holds a disjoint subset of the $p$ available features, $\mathcal{P}_1, \ldots, \mathcal{P}_K$ of size[2] $\tau = p/K$ belonging to the same observations. Throughout, we assume that the columns of $\mathbf{X}$ are normalized to have mean zero and unit variance. Let $\mathbf{X}_k \in \mathbb{R}^{n \times \tau}$ be the sub-matrix whose columns correspond to the coordinates in $\mathcal{P}_k$. The set $\mathcal{P}_{-k}$ contains all coordinates not in $\mathcal{P}_k$.

Each party aims to estimate $\boldsymbol{\beta}_k \in \mathbb{R}^\tau$, the portion of the true underlying parameter vector $\boldsymbol{\beta}$ corresponding to the features it holds, *while accounting for the contribution of the features held by the remaining parties*. However, due to privacy concerns the parties are not allowed to share their locally-held features. This scenario is of particular interest in healthcare and biomedicine (Que et al., 2012; Ohno-Machado, 2012; Li et al., 2015; Wu et al., 2012) but also in customer profiling and personalization.

**Example scenarios.** Here we briefly outline (non-exhaustively) two special cases of the general problem setting which cover a wide range of possible use cases—in particular in medical analyses—where a thorough accounting of confounding factors requires a mixture of public and private data to be aggregated.

(**A**) The data held by some parties is sensitive while the data held by other parties is public and not sensitive (e.g. Burkhardt et al. (2015)). It is possible to publish coefficients of the public blocks while still accounting for possible confounding effects of the private blocks.

---

2. For simplicity of notation only, in general the partitions can be of different sizes.

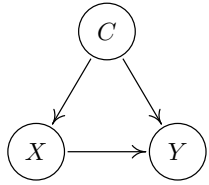

Figure 2: **Scenario (A)**. The confounders in $C$ influence both the variables in $X$ as well as the response $Y$.

(**B**) The response is not known to all parties. Then coefficients are only estimated for the blocks which know the response. The remaining parties just provide their data in secure form.

A concrete example of (**A**) is described graphically in Figure 2. Consider the response $Y$ being a variable measuring a cancer patient's health. Both genomic factors (contained in the set $C$) as well as gene expressions (in set $X$) have an influence on $Y$. In turn, genomic factors affect gene expressions. It is impossible to conduct a randomized study to estimate the effect of $X$ on $Y$ because gene expressions cannot be randomized. Additionally, due to its highly personal and sensitive nature, genomic data is rarely publicly available so $C$ and $X$ are stored separately (i.e. the full design matrix is vertically partitioned as in Burkhardt et al. (2015)). Due to the confounding links between $C$ and $X$, only including gene expressions in the model can result in heavily biased estimates for the effect of $X$ on $Y$ (Pearl, 2009).

Conducting studies that offer a holistic view on the factors influencing the response—as opposed to relying on biased estimates resulting from marginal studies—is tremendously important. However, it is an open question how to estimate the full model while providing formal privacy guarantees on the data sharing mechanism.

A further challenge comes from the observation that algorithms which preserve differential privacy typically scale poorly with dimensionality (Chaudhuri et al., 2011). Unfortunately, high-dimensional data is often encountered in biomedical applications where it can often be used to uniquely identify individuals due to the small sample sizes. This makes the need for privacy-aware algorithms which scale to high-dimensional problems more pressing.

## 3. The PRIDE algorithm

In this section we propose PRIDE, a scalable low-communication algorithm which extends the LOCO framework (Heinze et al., 2016) for distributed estimation to the *private* setting. Key to the PRIDE algorithm is the data sharing mechanism. The schematic is given in Figure 3. The full procedure is presented in Algorithm 1. We explain the following steps in more detail:

In **Step 2**, we compute the random features $(\mathbf{X}_k \mathbf{\Pi}_k) \in \mathbb{R}^{n \times \tau_{subs}}$. $\mathbf{\Pi}_k \in \mathbb{R}^{\tau \times \tau_{subs}}$ is the subsampled randomized Hadamard transform (SRHT) matrix which admits fast matrix-vector products (Tropp, 2010). We then perturb this by a Gaussian random matrix $\mathbf{W}_k \in$ $\mathbb{R}^{n \times \tau_{subs}} \sim \mathcal{N}(0, \sigma_k^2 \mathbf{I})$ to get $\widehat{\mathbf{Z}}_k = \mathbf{X}_k \mathbf{\Pi}_k + \mathbf{W}_k$. The exact form of $\sigma_k$ is given in Theorem 1.

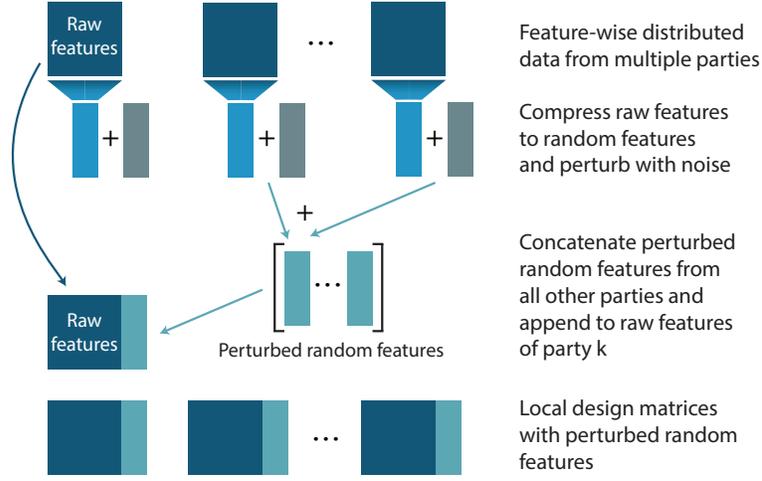

Figure 3: **Distributed differentially private data sharing mechanism used by PRIDE.**

In **Step 4**, the matrices of random features are communicated. For ease of notation, let $\tau_K = (K-1)\tau_{subs}$. Party $k$ then constructs the matrix

$$\bar{\mathbf{X}}_k \in \mathbb{R}^{n \times (\tau + \tau_K)} = \left[ \mathbf{X}_k, \left[ \widehat{\mathbf{Z}}_k \right]_{k' \neq k} \right], \qquad (2)$$

which is the column-wise concatenation of party $k$'s raw features and the perturbed random features from all other parties.

In **Step 5** each party solves the following local dual optimization problem

$$\tilde{\boldsymbol{\alpha}}_k = \underset{\boldsymbol{\alpha} \in \mathbb{R}^n}{\operatorname{argmax}} - \sum_{i=1}^n f_i^*(\alpha_i) - \frac{1}{2n\lambda} \boldsymbol{\alpha}^\top \bar{\mathbf{X}}_k \bar{\mathbf{X}}_k^\top \boldsymbol{\alpha}, \qquad (3)$$

where $f^*$ is the conjugate Fenchel dual of $f$. For example, for squared loss functions $f_i(u) = \frac{1}{2}(y_i - u)^2$, we have $f_i^*(\alpha) = \frac{1}{2}\alpha^2 + \alpha y_i$. This is solved using e.g. SDCA (Shalev-Shwartz and Zhang, 2013).

The main difference to DUAL-LOCO is the perturbation of the random features in **Step 2**. Although a small algorithmic difference, this has important consequences for the analysis which we present in the following section.

## 4. Analysis

For the discussion which follows we use the following definition of privacy which is concerned with changes in the attribute values of the observations rather than the difference in observations.

**Algorithm 1** PRIDE
**Input:** $Y$, $\mathbf{X}$ vertically-partitioned over $K$ parties, $\tau_{subs}, \lambda, \epsilon, \delta$

1: **for each** party $k \in \{1, \ldots K\}$ **in parallel do**
2:     Compute perturbed random features
       $\widehat{\mathbf{Z}}_k = \mathbf{X}_k \mathbf{\Pi}_k + \mathbf{W}_k$.
3:     Communicate $\widehat{\mathbf{Z}}_k$ to all parties $k'$ where $k' \neq k$.
4:     Construct local design matrix $\bar{\mathbf{X}}_k$.
5:     $\tilde{\boldsymbol{\alpha}}_k \leftarrow \texttt{LocalDualSolver}(\bar{\mathbf{X}}_k, Y, \lambda)$
6:     $\widehat{\boldsymbol{\beta}}_k = -\frac{1}{n\lambda} \mathbf{X}_k^\top \tilde{\boldsymbol{\alpha}}_k$
7: **end for**

**Output:** Each party $k$ obtains $\widehat{\boldsymbol{\beta}}_k$.

**Definition 1** (($\epsilon, \delta, \mathcal{S}$)-**differential privacy**) *A randomized algorithm* ALG *satisfies* ($\epsilon, \delta, \mathcal{S}$)-*differential privacy, if for all inputs* $\mathbf{X}$ *and* $\mathbf{X}'$ *differing in at most one user's one attribute value of an attribute in* $\mathcal{S} \subseteq \{1, \ldots, p\}$, *and for all sets of possible outputs* $\mathcal{D} \subseteq range(\text{ALG})$

$$\mathbb{P}\left[\text{ALG}(\mathbf{X}) \in \mathcal{D}\right] \leq e^\epsilon \, \mathbb{P}\left[\text{ALG}(\mathbf{X}') \in \mathcal{D}\right] + \delta \qquad (4)$$

*where the probability is computed over the randomness of the algorithm.*

When $\mathcal{S} = \{1, \ldots, p\}$, ($\epsilon, \delta, \mathcal{S}$)-differential privacy reduces to ($\epsilon, \delta$)-differential privacy. Informally, this states that (up to the parameters of the differential privacy guarantee) an adversary cannot infer a single attribute value for a single observation of an attribute in $\mathcal{S}$ from the output of the algorithm *despite knowing the values of all other attributes for all other observations.*

In the following definition, we use Definition 1 to formulate differential privacy in the distributed setting. The definition is close to Definition 2.4 in Beimel et al. (2011); here, we state it in our notation and for the case when $\delta > 0$.

**Definition 2** (($\epsilon, \delta$)-**distributed differential privacy**) *A randomized algorithm* ALG *satisfies* ($\epsilon, \delta$)-*distributed differential privacy, if* ALG *satisfies* ($\epsilon, \delta, \mathcal{S}$)-*differential privacy for all* $\mathcal{S} \in \{\mathcal{P}_{-k}; k = 1, \ldots, K\}$ *where* $\mathcal{P}_{-k}$ *is the set of indices corresponding to the features non-local to party* $k$.

A randomized algorithm ALG is ($\epsilon, 0$)-distributed differentially private if Definition 2.4 in Beimel et al. (2011) is fulfilled for $t = \max_k |\mathcal{P}_k|$. The condition in Beimel et al. (2011) is a bit stricter than ours as it requires ($\epsilon, \delta, \mathcal{S}$)-differential privacy for all sets $\mathcal{S}$ with $|\mathcal{S}^c| \leq t$ and not just for $\mathcal{P}_{-k}$ with $k = 1, \ldots, K$ as we do here. We also want to allow for $\delta > 0$ with Definition 2.

PRIDE achieves ($\epsilon, \delta$)-distributed differential privacy by perturbing random features with Gaussian noise before communicating them. As detailed in §4.1, this procedure preserves differential privacy according to Definition 1. While perturbing the random features has an adverse effect on the accuracy of the coefficient estimates, we prove an upper bound on the coefficient estimation error in §4.2. The error bound shows an interesting trade-off between the desired level of privacy and the accuracy of the random feature representation.

### 4.1 Distributed privacy guarantee

**Theorem 1 (Adapted from Kenthapadi et al. (2013))** *Let $w_2(\mathbf{\Pi}_k)$ denote the $\ell_2$-sensitivity of the projection matrix $\mathbf{\Pi}_k$ and let the range of the columns of $\mathbf{X}_k$ be bounded by $\theta_k$. $\mathcal{P}_{-k}$ is the set of indices corresponding to the features non-local to party $k$. Assuming $\delta < \frac{1}{2}$, let the entries of party $k$'s noise matrix $\mathbf{W}_k$ be drawn from $\mathcal{N}(0, \sigma_k^2 \mathbf{I})$ with*

$$\sigma_k > \frac{w_2(\mathbf{\Pi}_k) \cdot \theta_k}{\epsilon} \sqrt{2(\ln(1/2\delta) + \epsilon)}.$$

*Then* PRIDE *satisfies $(\epsilon, \delta)$-distributed differential privacy.*

The proof follows by adapting Kenthapadi et al. (2013) to hold for $(\epsilon, \delta)$-distributed differential privacy. When $\mathbf{\Pi}_k$ is the SRHT, $w_2(\mathbf{\Pi}_k) = 1$. Theorem 1 guarantees that an adversary who has access to the data held by party $k$ and knows all values of all attributes for every individual except for a single non-locally stored attribute value cannot infer that value from the perturbed random features which have been communicated to party $k$. This ensures PRIDE fulfills Definition 2. In contrast to the Laplace mechanism, the use of the Gaussian mechanism has the advantage that the required noise level is independent of the dimension of the projection matrix.

### 4.2 Approximation error of PRIDE

We now bound the coefficient approximation error between the PRIDE solution and the optimal solution to Eq. (1).

**Assumption 1** *Letting $r$ denote the rank of $\mathbf{X}$ and $\tau_K = (K-1)\tau_{subs}$, we require the following conditions to hold:*
  *A1. The projection dimension is chosen such that $\tau_K \gtrsim r \log r$.*
  *A2. The problem is high-dimensional, i.e. $n \leq p$, and $r = n$.*

**Theorem 2 (PRIDE approximation guarantee)** *Assume all $f_i$ in Eq. (1) to be convex and smooth with Lipschitz continuous gradients. Under Assumption 1 the overall error between the optimal solution to Eq. (1) $\boldsymbol{\beta}^*$ and the solution returned by PRIDE $\widehat{\boldsymbol{\beta}}$ is bounded with probability at least $1 - K\zeta$ by*

$$\|\widehat{\boldsymbol{\beta}} - \boldsymbol{\beta}^*\|_2 \leq \underbrace{\frac{\sqrt{K}\rho}{(1-2\rho)}\|\boldsymbol{\beta}^*\|_2}_{\text{(i)}} + \underbrace{\frac{\sqrt{K}\rho}{(1-2\rho)}\frac{\sigma}{d_{min}}\left(2 + \frac{\sigma\tau_K + \sigma\tau_K^2}{d_{min}}\right)\|\boldsymbol{\beta}^*\|_2}_{\text{(ii)}} \qquad (5)$$

*where $\rho = C\sqrt{\frac{r \log(2r/\xi)}{\tau_K}}$, $\sigma = \max_k \sigma_k$ and $d_{min} = d_r(\mathbf{X})$, the smallest non-zero singular value of $\mathbf{X}$. $C$ and $\xi$ are absolute positive constants. The exact form of $\zeta$ is given in §A.*

*Proof strategy.* (Full details are given in §A.) We require to bound the local coefficient estimation error of a single party $k$ which can then be combined with a union bound to obtain the global approximation error. To bound the local error (Theorem 3), a key step is bounding the difference between the full (non-perturbed, single-party) kernel matrix $\mathbf{K}$ and the projected-and-perturbed kernel matrix $\tilde{\mathbf{K}}$ (omitting the subscript $k$ for ease of notation) where

$$\mathbf{K} = \mathbf{X}\mathbf{X}^\top \quad \text{and} \quad \tilde{\mathbf{K}} = (\mathbf{X}\Theta + \mathbf{E})(\mathbf{X}\Theta + \mathbf{E})^\top,$$

$$\Theta = \text{diag}(\mathbf{I}_\tau, \mathbf{\Pi}_1, \ldots, \mathbf{\Pi}_{K-1}) \in \mathbb{R}^{p \times (\tau + \tau_K)} \quad \text{and}$$

$$\mathbf{E} = \begin{bmatrix} \mathbf{0}_\tau & \mathbf{W}_1 & \ldots & \mathbf{W}_{K-1} \end{bmatrix} \in \mathbb{R}^{n \times (\tau + \tau_K)}.$$

When privacy is not required, $\sigma = 0$ and $\mathbf{E} = \mathbf{0}$ in which case we recover the approximation guarantee of DUAL-LOCO which relies on the fact that $\|\mathbf{K} - \tilde{\mathbf{K}}\|_2 \leq \rho$ (Heinze et al., 2016). However, this bound does not hold when i.i.d. Gaussian noise is added to those entries of $\mathbf{X}\Theta$ corresponding to the random features (i.e. $\sigma > 0$ and $\mathbf{E} \neq \mathbf{0}$). Now, we require to find an upper bound on $\|\mathbf{K} - \tilde{\mathbf{K}}\|_2$ and the proof also requires a lower bound on $\|\tilde{\mathbf{K}}\|_2$. We can bound $\|\mathbf{K} - (\mathbf{X}\Theta)(\mathbf{X}\Theta)^\top\|_2 \leq \rho$ and use Lemma 6 to bound the terms involving $\mathbf{E}$ with high probability. While the exact expressions are more involved, intuitively, in expectation the cross terms are zero while the diagonal elements of $\mathbf{E}\mathbf{E}^\top$ are at most $\sigma^2 \tau_K$.

Finally, lower bounding $\tilde{\mathbf{K}}$ requires a different technique as the involved cross terms are not positive semidefinite. Using that terms involving $\mathbf{E}$ are centered around 0 and applying a Chernoff bound (Lemma 7) allows us to show $\|\tilde{\mathbf{K}}\|_2 \geq 1 - 2\rho$. Full details are given in §A. $\square$

**Discussion.** The bound in Theorem 2 consists of two terms: **(i)** The approximation error due to the (distributed) random projection representation. This decreases as the projection dimension $\tau_{subs}$ increases, providing a more accurate approximation to the non-local features. **(ii)** The error due to the perturbation necessary for guaranteeing privacy. This term is increasing in $\tau_{subs}$— a larger dimensional random feature representation contributes more noisy dimensions which act like an additional $\ell_2$-regularizer. This can be seen clearly when comparing the solutions to the dual formulation of the ridge regression objective: The optimal solution is given by $\boldsymbol{\alpha}^* = (\mathbf{K} + \lambda \mathbf{I})^{-1} Y$ while party $k$ computes $\tilde{\boldsymbol{\alpha}} = ((\mathbf{X}\Theta + \mathbf{E})(\mathbf{X}\Theta + \mathbf{E})^\top + \lambda \mathbf{I})^{-1} Y$. The diagonal elements of $\mathbf{E}\mathbf{E}^\top$ are centered around $\sigma^2 \tau_K = \sigma^2(K-1)\tau_{subs}$, so using a larger projection dimension $\tau_{subs}$ increases the regularizing effect (and therefore bias) induced by $\mathbf{E}\mathbf{E}^\top$ which acts in addition to the one caused by $\lambda$.[3] On the other hand, the bias can be decreased by increasing $\epsilon$ (decreasing $\sigma$) implying a weaker privacy guarantee.

We thus observe a trade-off between approximation quality and privacy. When a very strong privacy guarantee is required—implying a large value of $\sigma$—a smaller $\tau_{subs}$ should be chosen so that the additional regularization does not become too strong. On the other hand, if the privacy requirements are less stringent, a larger $\tau_{subs}$ together with a larger $\epsilon$ will yield better approximation quality. In general, PRIDE will be most effective when the rank of the problem is such that a relatively small projection dimension will capture most of the important structure in the data. We demonstrate the effect of this trade-off empirically in the following section. Importantly, we shall see that the induced bias that results from not communicating any data is often much larger than the bias of the PRIDE estimates.

---

3. In §C we show that for the primal formulation of the least-squares objective, the effect of $\mathbf{E}$ can be understood as an $\ell_2$-regularizer which acts on the random features only.

Table 1: **Data set statistics**. $\max_k (\theta_k)$ is the largest bound on the range of the columns of $\mathbf{X}_k$ among all parties, $d_{\min}$ is the smallest non-zero singular value of the design matrix, $r_{\text{eff}}(\mathbf{X})$ denotes the effective rank and $J_u$ denotes the number of principal components that capture $u\%$ of the variance in the data set.

|  | $n_{\text{train}}$ | $p$ | $\max_k (\theta_k)$ | $d_{\min}$ | $r_{\text{eff}}(\mathbf{X})$ | $J_{80}$ | $J_{90}$ |
|---|---|---|---|---|---|---|---|
| SIM. | 800 | 400 | 7.41 | 3.7e-6 | 2.03 | 3 | 5 |
| CLIMATE | 849 | 10,368 | 8.51 | 3.32 | 4.03 | 29 | 54 |
| CANCER | 188 | 2,000 | 10.92 | 11.57 | 4.53 | 65 | 107 |

## 5. Experiments

We present results on three datasets summarized in Table 1. Also reported are the smallest non-zero singular value of the design matrix, $d_{\min}$; the effective rank $r_{\text{eff}}(\mathbf{X}) = \text{tr}\left(\mathbf{X}^\top \mathbf{X}\right)/\|\mathbf{X}\|_2^2$ and the largest bound on the range of the columns of $\mathbf{X}_k$ among all parties, $\max_k (\theta_k)$. The effective rank (Vershynin, 2010) is a measure of the intrinsic dimension of a matrix which captures whether the matrix lies near to a low-dimensional subspace.

We compare the performance of five methods:

- "Semi-Naive Bayes" (NB). Here, a separate model is learned by each party independently:

$$\widehat{\boldsymbol{\beta}}_k^{\text{NB}} = \text{argmin}_{\mathbf{b}_k} \sum_{i=1}^n f_i(\mathbf{x}_k^\top \mathbf{b}_k) + \lambda \|\mathbf{b}_k\|_2^2.$$

  Since no data is communicated, the features are kept completely private.

- The standard DUAL-LOCO algorithm (corresponding to PRIDE with $\sigma_k = 0 \ \forall k$). Since the random features are not perturbed this does not guarantee privacy according to Theorem 1.

- Our proposed PRIDE algorithm. We show the effect of varying the privacy parameter $\epsilon$ by varying the noise variance $\sigma_k^2$. We fix $\delta = 0.05$ as varying $\delta$ has only little effect on $\sigma_k$. As $\sigma_k$ also depends on the maximal range of the columns of $\mathbf{X}_k$, we report the maximum of $\sigma_k$ for $k = 1, \ldots, K$ in Table D.2.

- In the non-distributed setting: GLMNET (Friedman et al., 2010) and SDCA (Shalev-Shwartz and Zhang, 2013).

For both DUAL-LOCO and PRIDE we show results for different values of the projection dimension $\tau_{subs}$. The absolute dimensions are given in Table D.1. Details on the cross validation procedure are given in §E.

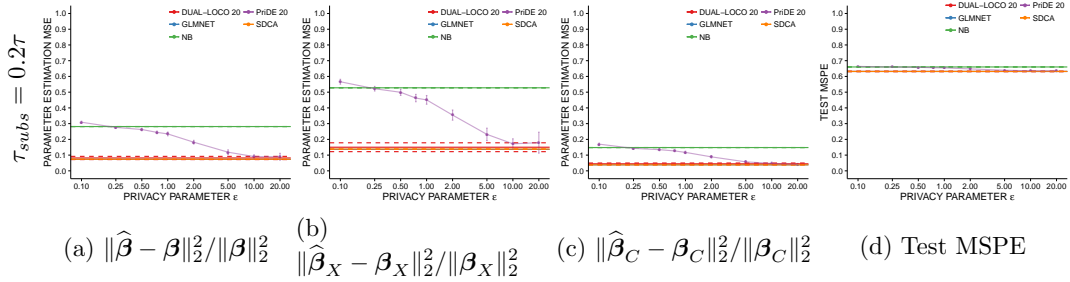

(a) $\|\widehat{\boldsymbol{\beta}} - \boldsymbol{\beta}\|_2^2/\|\boldsymbol{\beta}\|_2^2$ (b) $\|\widehat{\boldsymbol{\beta}}_X - \boldsymbol{\beta}_X\|_2^2/\|\boldsymbol{\beta}_X\|_2^2$ (c) $\|\widehat{\boldsymbol{\beta}}_C - \boldsymbol{\beta}_C\|_2^2/\|\boldsymbol{\beta}_C\|_2^2$ (d) Test MSPE

Figure 4: **Simulated data**. Results for projection dimension $\tau_{subs} = 0.2\tau$. Parameter estimation errors are computed w.r.t. the data generating model. Additional plots for $\tau_{subs} = \{0.05, 0.1\} \cdot \tau$ can be found in Figure D.2 in the supplementary information.

### 5.1 Simulated data

We revisit example (**A**) given in §2. The data are simulated according to the model in Figure 2.[4] We consider two blocks of features, $C$ and $X$. For example, $C$ could contain genomic data such as measurements of single nucleotide polymorphisms (SNPs). Due to its highly personal and sensitive nature, genomic data arising from techniques like SNP genotyping is rarely publicly available. The other block, $X$ could hold gene expression data. Some of the genomic features have an effect on some of the gene expression features and both sets of features contribute to the response $Y$.

We distribute the two blocks of features over $K = 2$ parties so that $X$ and $C$ are kept separately. In this experiment we aim to analyze the parameter estimation error with respect to the *true underlying coefficients* $\boldsymbol{\beta}$. Due to the dependence between $C$ and $X$ one cannot obtain accurate coefficient estimates for the effect of $X$ on $Y$ when only including $X$ into the model. We aim to assess whether the perturbed random projections used by PriDE suffice to communicate enough information to obtain accurate estimates in this challenging estimation task.

Comparisons of normalized coefficient estimation error with respect to the data generating coefficients $\boldsymbol{\beta}$ are shown for $\tau_{subs} = 0.2\tau$ in Figure 4a-c. There is a significant difference between the NB and Dual-Loco and SDCA solutions, particularly for block $X$. This performance gap is to be expected due to the confounding effect of $C$. It shows that in order to obtain accurate coefficient estimates in the distributed setting some degree of communication is crucial which allows to adjust for the dependencies between the features. For small $\epsilon$ (more privacy) PriDE performs similarly to NB, i.e. the incurred biases are on the same scale. As $\epsilon$ increases, PriDE approaches and eventually equals the performance of Dual-Loco and SDCA. This demonstrates that PriDE is able to approximate the true $\boldsymbol{\beta}$ accurately for sufficiently large values of $\epsilon$. Thus PriDE allows to adjust for the confounding effects from $C$ on $X$ while guaranteeing $(\epsilon, \delta, \mathcal{S})$-differential privacy.

Figure 4d shows the normalized prediction MSE on the test set. All methods perform similarly. Due to the confounding effect of $C$ and $X$, NB is unable to obtain accurate coefficient estimates but it can achieve good predictive performance in this example.

---
4. Full simulation details are given in §D.1 and the data generating code is provided as a supplement to this work.

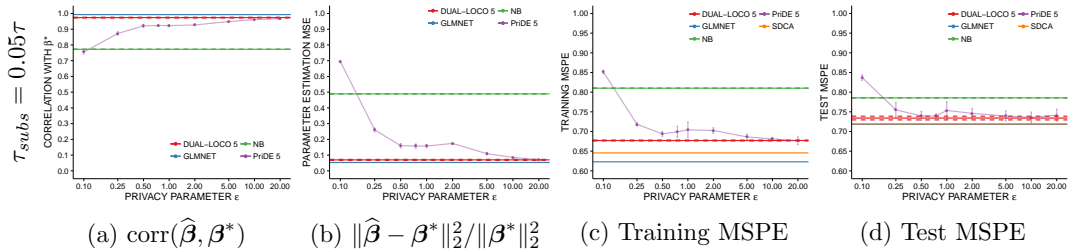

Figure 5: **Climate model data**. Results for projection dimension $\tau_{subs} = 0.05\tau$. The parameter estimation metrics are computed w.r.t. the optimal single-machine solution $\boldsymbol{\beta}^*$ obtained with SDCA. Additional plots for $\tau_{subs} = \{0.01, 0.1, 0.2\} \cdot \tau$ can be found in Figure D.3.

This experiment also suggests that Assumption 1 can be weakened to settings where the effective rank of the data is low while $n > p$. Different proof techniques would be required to extend Thereom 2 to such cases.

### 5.2 Climate model data

Next, we present an application to a problem in climate modeling. We consider data from part of the CMIP5 climate modeling ensemble which are taken from control simulations of the GISS global circulation model (Schmidt et al., 2014). We aim to forecast the monthly global average temperature $Y$ in February using the air pressure measured in January. The features are pressure measurements taken at $p = 10,368$ geographic grid points. The model simulates the climate for a range of 531 years and we use the output from two control simulation runs. The data set is split into training ($n_{\text{train}} = 849$) and test set ($n_{\text{test}} = 213$), and we distribute the problem across $K = 4$ parties.

Comparisons of correlation and estimation error, global training and test error (all normalized) are shown for $\tau_{subs} = 0.05\tau$ in Figure 5. Since the true coefficients $\boldsymbol{\beta}$ are unknown in this example, comparisons of correlation and estimation error are computed with respect to the empirical risk minimizer, i.e. the optimal parameters $\boldsymbol{\beta}^*$, estimated using SDCA where all the data was available, non-perturbed, on a single machine. The parameter estimation error is the quantity which is bounded in Theorem 2.

Importantly, there is a significant difference between the NB and DUAL-LOCO solutions. This performance gap shows that in the distributed setting some degree of communication is crucial for good statistical estimation and predictive accuracy for this problem as not communicating any features incurs a large bias. In Figure D.3 we observe that increasing $\tau_{subs}$ does not cause a large change in the accuracy achieved by DUAL-LOCO. This suggests that the problem is nominally low rank and a small projection dimension suffices to capture the structure of the data. This is to be expected given the high degree of spatial correlation of pressure measurements and is confirmed by the estimate of the effective rank and the PCA statistics in Table 1. For reasonable values of $\epsilon$, the PRIDE solution quickly approaches the DUAL-LOCO solution for all four measures of accuracy. Importantly, PRIDE achieves a test prediction error within the margin of error of the DUAL-LOCO prediction error.

We observe the trade-off implied by Theorem 2: As the projection dimension increases, the DUAL-LOCO approximation error decreases (i.e. term (i) in Eq. (5)). However, term (ii) grows with $\sigma^2(K-1)\tau_{subs}$. For very small values of $\epsilon$ this second contribution dominates so a smaller projection dimension typically yields better performance. As $\epsilon$ increases, the gain in approximation quality starts to outweigh the regularization bias incurred by increasing $\tau_{subs}$, so that for large values of $\epsilon$ a large projection dimension performs best. In Figure D.3, this trade-off is reflected in a slower convergence to the DUAL-LOCO solution for larger values of $\tau_{subs}$.

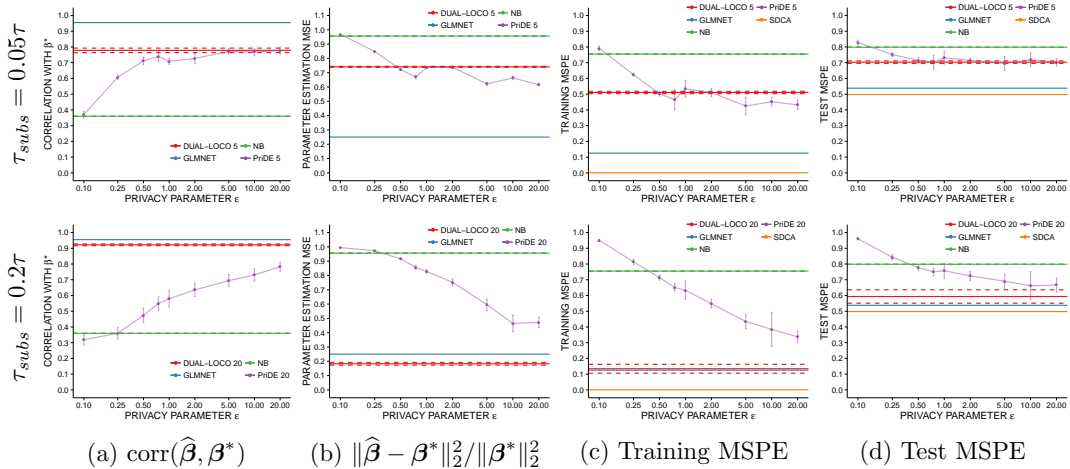

Figure 6: **Gene expression data**. Results for projection dimension $\tau_{subs} = \{0.05, 0.2\} \cdot \tau$. Further results for $\tau_{subs} = \{0.01, 0.1\} \cdot \tau$ can be found in Figure D.4.

### 5.3 Breast cancer gene expression data

Finally, we show an application to a problem in clinical bioinformatics. This experiment aims to assess the performance of PRIDE on a real data set from a domain where sensitive data is ubiquitous. We use the breast cancer data set GSE3494[5] (Miller et al., 2005). Our task is to predict the disease specific survival time of each patient in years where the objective can be reformulated via an Accelerated Failure Time model as a least squares objective; here with constant weights. The median follow-up of patients was 122 months. Approximately $45{,}000$ gene expressions are available from $n = 236$ patients. We selected genes with the largest absolute marginal correlation with the response, resulting in $p = 2{,}000$ distributed across $K = 4$ parties. The data set is split into training ($n_{\text{train}} = 188$) and test set ($n_{\text{test}} = 48$). In this application, accurate estimates of $\boldsymbol{\beta}$ are of primary interest to assess which genes are good predictors for survival time.

Figure 6a and b show comparisons of correlation and estimation error for $\tau_{subs} = \{0.05, 0.2\} \cdot \tau$ with respect to the SDCA solution $\boldsymbol{\beta}^*$. Columns c and d show the normalized training and test prediction MSE. We again observe a large difference between the NB and DUAL-LOCO solutions which is essential for there to be some expected gain from using PRIDE. We observe similar trends as in the previous experiments: as $\epsilon$ increases, PRIDE improves upon NB and approaches the DUAL-LOCO solution. The trade-off be-tween $\epsilon$ and $\tau_{subs}$ (implied by Theorem 2 and discussed above) is again apparent. The convergence to the DUAL-LOCO solution with increasing values of $\epsilon$ is somewhat slower than in the previous experiment. This is partly due to the gene expression data having heavier tails, resulting in a larger $\max_k (\theta_k)$. This requires a larger noise level to guarantee privacy leading to a more heavily regularized learning problem (cf. Tables 1 and D.2).

---

5. http://www.ncbi.nlm.nih.gov/geo/query/acc.cgi?acc=GSE3494.

In summary, the behavior predicted by Theorem 2 is confirmed empirically. The best performance of PRIDE can be obtained by finding the optimum of the accuracy-privacy trade-off, respecting the problem-specific constraints on privacy. That is, by choosing a projection dimension $\tau_{subs}$ that suffices to capture the signal contained in the non-local features, so that term (i) in Eq. (5) is as small as possible without over-regularizing the objective and introducing a large bias from term (ii). Finding the optimal projection dimension is then a problem of model selection. We discuss the challenges of a privacy preserving cross validation scheme in §E.

In general, given a suitable projection dimension, PRIDE can significantly improve upon the NB solution: the bias of the NB estimates is often much larger than the bias of the PRIDE estimates. This suggests that the PRIDE framework allows for accurate distributed statistical estimation while guaranteeing $(\epsilon, \delta, \mathcal{S})$-differential privacy.

## 6. Related work

**Privacy-aware learning.** Ensuring differential privacy in supervised learning techniques has garnered increasing interest in recent years (Chaudhuri and Monteleoni, 2009; Chaudhuri et al., 2011) and approaches have been proposed to solve more general convex optimization problems in a private fashion (Song et al., 2013). These approaches achieve privacy by either applying noise to the coefficient vector before it is returned or perturbing the objective with noise during optimization.

Kenthapadi et al. (2013) apply a Johnson-Lindenstrauss random projection to compress the column space of the design matrix and perturb the resulting matrix with Gaussian noise. This procedure allows the compressed, perturbed data matrix to be published but forfeits the interpretability of the features as any subsequent queries must be performed in the compressed space. This approach is related to *local privacy* (Duchi et al., 2013) where the algorithm only observes a disguised version of the data.

**Distributed statistical estimation.** Distributed estimation and optimization when the data is horizontally partitioned has been a popular topic in recent years (Jaggi et al., 2014; Zhang et al., 2013). However, the problem of statistical estimation when the data is vertically partitioned has been less well studied since most loss functions of interest are not separable across coordinates. A key challenge addressed by Heinze et al. (2014, 2016) was to define a local minimization problem for each worker to solve *independently* while still maintaining important dependencies between features held by different parties. This is achieved by communicating low-dimensional random projections of the data held by each party which keeps communication overhead low. Although this obfuscates the data to some degree, it does not guarantee privacy.

**Preserving privacy in distributed learning.** McGregor et al. (2011) introduce the notion of two-party differential privacy which is generalized to an arbitrary number of parties in Beimel et al. (2011). We discuss the relation to Definitions 1 and 2 in §4.1.

In distributed supervised learning, private approaches have been much less studied and focus mainly on the setting where data is horizontally partitioned (Huang et al., 2015; Zhang and Zhu, 2016). Few approaches have been considered for privacy preserving learning in the distributed setting when the data is partitioned *vertically* (Yu et al., 2006; Mangasarian et al., 2008; Wu et al., 2012; Mohammed et al., 2014). However, no formal guarantees with respect to both privacy and utility are given.

## 7. Conclusions and further work

We have proposed PriDE which addresses some of the important concerns in learning from sensitive, vertically partitioned data in a principled and scalable way. PriDE preserves $(\epsilon, \delta)$-distributed differential privacy while maintaining a low approximation error with respect to the optimal, non-private, non-distributed model. To the best of our knowledge, no other methods with similar guarantees have been proposed for the considered problem setting.

PriDE only communicates perturbed low-dimensional random projections of the original features so the communication overhead is small. Since estimation is performed on a combination of raw and random features, the solution is returned in the original space preserving interpretability of the coefficients. This allows to assess a feature's impact on the response while accounting for the contribution of—possibly confounding—sensitive features held by other parties. For prediction tasks, each party can use its own local design matrix, consisting of raw and perturbed random features.

Empirically, we have shown on simulated and real-world datasets that the PriDE estimates greatly improve upon the performance of the fully-private semi-Naive Bayes model and approach (i) the true underlying coefficients, and (ii) the estimates of the non-private and non-distributed Glmnet and SDCA solvers. PriDE also performs well in areas not specifically covered by Theorem 2, as shown for the low-dimensional synthetic data. This suggests that our result could be generalized to when the effective rank of the problem is small.

Perturbing the random features is necessary to preserve privacy but adds bias to the solution. An open question concerns whether recent approaches to errors-in-variables regression (Loh and Wainwright, 2012) could be used to obtain an unbiased solution and perhaps improve the performance of local CV. For ensuring differentially private public coefficient release, existing techniques such as perturbing the coefficients, objective (Chaudhuri and Monteleoni, 2009; Chaudhuri et al., 2011), or dual variables (Zhang and Zhu, 2016) with additive heavy-tailed noise may be used.

Po-Ling Loh and Martin J Wainwright. High-dimensional regression with noisy and missing data: Provable guarantees with nonconvexity. *The Annals of Statistics*, 40(3):1637–1664, June 2012.

Olvi L Mangasarian, Edward W Wild, and Glenn M Fung. Privacy-preserving classification of vertically partitioned data via random kernels. *ACM TKDD*, 2(3):12, 2008.

Andrew McGregor, Ilya Mironov, Toniann Pitassi, Omer Reingold, Kunal Talwar, and Salil P. Vadhan. The limits of two-party differential privacy. *Electronic Colloquium on Computational Complexity (ECCC)*, 18:106, 2011.

Lance D Miller, Johanna Smeds, Joshy George, and Vinsensius Band others Vega. An expression signature for p53 status in human breast cancer predicts mutation status, transcriptional effects, and patient survival. *Proceedings of the National Academy of Sciences of the United States of America*, 102(38):13550–13555, 2005.

Noman Mohammed, Dima Alhadidi, Benjamin C. M. Fung, and Mourad Debbabi. Secure two-party differentially private data release for vertically partitioned data. *IEEE Trans. Dependable Sec. Comput.*, 11(1):59–71, 2014.

Lucila Ohno-Machado. To share or not to share: That is not the question. *Science Translational Medicine*, 4(165), 2012. ISSN 1946-6234.

Judea Pearl. *Causality: Models, Reasoning, and Inference*. Cambridge University Press, 2nd edition, 2009.

Jialan Que, Xiaoqian Jiang, and Lucila Ohno-Machado. A collaborative framework for distributed privacy-preserving support vector machine learning. In *AMIA Annual Symposium Proceedings*, 2012.

Anand D Sarwate, Sergey M Plis, Jessica A Turner, Mohammad R Arbabshirani, and Vince D Calhoun. Sharing privacy-sensitive access to neuroimaging and genetics data: a review and preliminary validation. *Frontiers in neuroinformatics*, 8, 2014.

Gavin A. Schmidt, Max Kelley, Larissa Nazarenko, Reto Ruedy, et al. Configuration and assessment of the GISS ModelE2 contributions to the CMIP5 archive. *Journal of Advances in Modeling Earth Systems*, 6(1):141–184, 2014.

Shai Shalev-Shwartz and Tong Zhang. Stochastic dual coordinate ascent methods for regularized loss. *The Journal of Machine Learning Research*, 14(1):567–599, 2013.

Shuang Song, Kamalika Chaudhuri, and Anand D Sarwate. Stochastic gradient descent with differentially private updates. In *GlobalSIP*, pages 245–248. IEEE, 2013.

Joel A. Tropp. Improved analysis of the subsampled randomized Hadamard transform. November 2010. arXiv:1011.1595v4 [math.NA].

Roman Vershynin. Introduction to the non-asymptotic analysis of random matrices. November 2010. arXiv:1011.3027.

Yuan Wu, Xiaoqian Jiang, Jihoon Kim, and Lucila Ohno-Machado. Grid binary logistic regression (glore): building shared models without sharing data. *Journal of the American Medical Informatics Association*, 19(5):758–764, 2012.

Hwanjo Yu, Jaideep Vaidya, and Xiaoqian Jiang. Privacy-preserving svm classification on vertically partitioned data. In *Advances in Knowledge Discovery and Data Mining*, pages 647–656. Springer, 2006.

# Supplementary Information

## Appendix A. Proofs

We first state and prove a theorem which bounds the estimation error for a single party, $k$. The proof of Theorem 2 follows straightforwardly from combining this result for all $K$ parties and applying a union bound.

**Theorem 3** *Assume all $f_i$ in Eq. (1) to be convex and smooth with Lipschitz continuous gradients. Under Assumption 1, the local difference between the optimal solution to Eq. (1) at party $k$, $\boldsymbol{\beta}_k^*$, and the solution returned by PRIDE at party $k$, $\widehat{\boldsymbol{\beta}}_k$, is bounded with probability $1 - \zeta$ by*

$$\|\widehat{\boldsymbol{\beta}}_k - \boldsymbol{\beta}_k^*\|_2 \leq \frac{\rho}{(1-2\rho)}\left(1 + \frac{\sigma}{d_{min}}\left(2 + \frac{\sigma\tau_K + \sigma\tau_K^2}{d_{min}}\right)\right)\|\boldsymbol{\beta}^*\|_2 \tag{6}$$

*where $\zeta = 3c_1 \exp(-c_2 \log r) + 2\xi + 2\frac{p}{e^r} + e^{-(\tau+\tau_K)/16}$, $\tau_K = (K-1)\tau_{subs}$, $\rho = C\sqrt{\frac{r\log(2r/\xi)}{\tau_K}}$, $\sigma = \max_k \sigma_k$ and $d_{min} = d_r(\mathbf{X})$, the smallest non-zero singular value of $\mathbf{X}$. $C, c_1, c_2,$ and $\xi$ are absolute positive constants.*

**Definition 3** *For ease of exposition, we shall rewrite the dual problems so that we consider minimizing convex objective functions. More formally, the original problem is then given by*

$$\boldsymbol{\alpha}^* = \underset{\boldsymbol{\alpha} \in \mathbb{R}^n}{\operatorname{argmin}}\left\{D(\boldsymbol{\alpha}) := \sum_{i=1}^n f_i^*(\alpha_i) + \frac{1}{2n\lambda}\boldsymbol{\alpha}^\top \mathbf{K}\boldsymbol{\alpha}\right\}. \tag{7}$$

*The problem party $k$ solves is described by*

$$\tilde{\boldsymbol{\alpha}} = \underset{\boldsymbol{\alpha} \in \mathbb{R}^n}{\operatorname{argmin}}\left\{\tilde{D}_k(\boldsymbol{\alpha}) := \sum_{i=1}^n f_i^*(\alpha_i) + \frac{1}{2n\lambda}\boldsymbol{\alpha}^\top \tilde{\mathbf{K}}_k \boldsymbol{\alpha}\right\}. \tag{8}$$

*Recall that $\tilde{\mathbf{K}}_k = \bar{\mathbf{X}}_k \bar{\mathbf{X}}_k^\top$, where $\bar{\mathbf{X}}_k$ is the concatenation of the $\tau$ raw features and $(K-1)\tau_{subs}$ perturbed random features for party $k$ as in **Step 4** of Algorithm 1.*

**Proof** [Proof of Theorem 3] For ease of notation, we shall omit the subscript $k$ in $\tilde{\mathbf{K}}_k$ in the following. Defining

$$\Theta = \begin{bmatrix} \mathbf{I}_\tau & 0 & \ldots & 0 \\ 0 & \mathbf{\Pi}_1 & 0 & \vdots \\ \vdots & \ldots & \ddots & 0 \\ 0 & \ldots & \ldots & \mathbf{\Pi}_{K-1} \end{bmatrix} \in \mathbb{R}^{p \times (\tau+(K-1)\tau_{subs})} \tag{9}$$

and

$$\mathbf{E} = \begin{bmatrix} \mathbf{0}_\tau & \mathbf{W}_1 & \ldots & \mathbf{W}_{K-1} \end{bmatrix} \in \mathbb{R}^{n \times (\tau+(K-1)\tau_{subs})}, \tag{10}$$

we can write the original as well as the projected and perturbed kernel matrices explicitly as

$$\mathbf{K} = \mathbf{X}\mathbf{X}^\top \qquad \text{and} \qquad \tilde{\mathbf{K}} = (\mathbf{X}\Theta + \mathbf{E})(\mathbf{X}\Theta + \mathbf{E})^\top$$

respectively. Applying Lemma 4, on the l.h.s. of (18) we have

$$(\tilde{\boldsymbol{\alpha}} - \boldsymbol{\alpha}^*)^\top (\mathbf{K} - \tilde{\mathbf{K}})\boldsymbol{\alpha}^* = (\tilde{\boldsymbol{\alpha}} - \boldsymbol{\alpha}^*)^\top \left( \mathbf{XX}^\top - (\mathbf{X}\boldsymbol{\Theta})(\mathbf{X}\boldsymbol{\Theta})^\top - (\mathbf{X}\boldsymbol{\Theta}\mathbf{E}^\top) - (\mathbf{X}\boldsymbol{\Theta}\mathbf{E}^\top)^\top - \mathbf{EE}^\top \right) \boldsymbol{\alpha}^*.$$

Denoting $\mathbf{UDV}^\top = \mathbf{X}$, $\tilde{\boldsymbol{\gamma}} = \mathbf{DU}^\top \tilde{\boldsymbol{\alpha}}$ and $\boldsymbol{\gamma}^* = \mathbf{DU}^\top \boldsymbol{\alpha}^*$ we have

$$(\tilde{\boldsymbol{\alpha}} - \boldsymbol{\alpha}^*)^\top (\mathbf{K} - \tilde{\mathbf{K}})\boldsymbol{\alpha}^*$$
$$= (\tilde{\boldsymbol{\gamma}} - \boldsymbol{\gamma}^*)^\top \left( \mathbf{V}^\top \mathbf{V} - \mathbf{V}^\top \boldsymbol{\Theta}\boldsymbol{\Theta}^\top \mathbf{V} - \mathbf{V}^\top \boldsymbol{\Theta} \mathbf{E}^\top \mathbf{U} \mathbf{D}^{-1} - \mathbf{D}^{-1} \mathbf{U}^\top \mathbf{E} \boldsymbol{\Theta}^\top \mathbf{V} - \mathbf{D}^{-1} \mathbf{U}^\top \mathbf{EE}^\top \mathbf{U} \mathbf{D}^{-1} \right) \boldsymbol{\gamma}^*.$$

By Assumption 1 $r = n$, so we have $\mathbf{UU}^\top = \mathbf{I}_n$. Adding and subtracting $(\tilde{\boldsymbol{\gamma}} - \boldsymbol{\gamma}^*)^\top \mathbf{D}^{-1} \mathbf{U}^\top \left( \sigma^2 (K-1)\tau_{subs} \mathbf{I} \right) \mathbf{UD}^{-1} \boldsymbol{\gamma}^*$ where $\sigma^2 = \max_k(\sigma_k^2)$ yields

$$(\tilde{\boldsymbol{\alpha}} - \boldsymbol{\alpha}^*)^\top (\mathbf{K} - \tilde{\mathbf{K}})\boldsymbol{\alpha}^* = (\tilde{\boldsymbol{\gamma}} - \boldsymbol{\gamma}^*)^\top \underbrace{\left( \mathbf{I}_r - \mathbf{V}^\top \boldsymbol{\Theta}\boldsymbol{\Theta}^\top \mathbf{V} \right)}_{(\mathrm{I})} \boldsymbol{\gamma}^*$$

$$- (\tilde{\boldsymbol{\gamma}} - \boldsymbol{\gamma}^*)^\top \underbrace{\left( \mathbf{V}^\top \boldsymbol{\Theta} \mathbf{E}^\top \mathbf{U} \mathbf{D}^{-1} + \mathbf{D}^{-1} \mathbf{U}^\top \mathbf{E} \boldsymbol{\Theta}^\top \mathbf{V} \right)}_{(\mathrm{II})} \boldsymbol{\gamma}^*$$

$$+ (\tilde{\boldsymbol{\gamma}} - \boldsymbol{\gamma}^*)^\top \underbrace{\mathbf{D}^{-1} \mathbf{U}^\top \left( \sigma^2(K-1)\tau_{subs} \mathbf{I} - \mathbf{EE}^\top \right) \mathbf{UD}^{-1}}_{(\mathrm{III})} \boldsymbol{\gamma}^*$$

$$- (\tilde{\boldsymbol{\gamma}} - \boldsymbol{\gamma}^*)^\top \underbrace{\mathbf{D}^{-1} \mathbf{U}^\top \left( \sigma^2(K-1)\tau_{subs} \mathbf{I} \right) \mathbf{UD}^{-1}}_{(\mathrm{IV})} \boldsymbol{\gamma}^*.$$

We now focus on bounding each of the terms **(I)**, **(II)**, **(III)** and **(IV)** in turn.

**(I).** This term is bounded with probability $1 - \left( \xi + \frac{p-\tau}{e^r} \right)$ by $\rho_1 = \sqrt{\frac{cr \log(2r/\xi)}{(K-1)\tau_{subs}}}$ which follows directly from applying Lemma 5.

**(II).** We aim to bound the term $\mathbf{D}^{-1} \mathbf{U}^\top \mathbf{E} \boldsymbol{\Theta}^\top \mathbf{V} \boldsymbol{\gamma}^*$. Since the random terms are sub-Gaussians, we will bound this term using Lemma 6 with $Y = (\mathbf{U}^\top \mathbf{E})^\top \in \mathbb{R}^{\tau + (K-1)\tau_{subs} \times r}$, $X = (\boldsymbol{\Theta}^\top \mathbf{V}) \in \mathbb{R}^{\tau + (K-1)\tau_{subs} \times r}$ and $\mathbb{E}\left[ \mathbf{U}^\top \mathbf{E} \boldsymbol{\Theta}^\top \mathbf{V} \right] = \mathbf{0}$. Since the first $\tau$ rows of $Y$ are $Y_0 = 0$, we decompose $Y$ and the corresponding rows in $X$ as

$$Y = \begin{bmatrix} Y_0 = \mathbf{0} \\ Y_1 \end{bmatrix} \qquad X = \begin{bmatrix} X_0 \\ X_1 \end{bmatrix}.$$

According to this decomposition we can write $Y^\top X = Y_0^\top X_0 + Y_1^\top X_1$. Clearly $Y_0^\top X_0 = 0$ so $Y^\top X$ only has $(K-1)\tau_{subs}$ non-zero summands. Now, applying Lemma 6, with probability $1 - c_1 \exp(-c_2 \log r)$

$$\|\mathbf{D}^{-1} \mathbf{U}^\top \mathbf{E} \boldsymbol{\Theta}^\top \mathbf{V} \boldsymbol{\gamma}^*\|_2 \leq \frac{1}{d_{\min}} \|\mathbf{U}^\top \mathbf{E} \boldsymbol{\Theta}^\top \mathbf{V} \boldsymbol{\gamma}^*\|_2$$

$$\leq \frac{\sqrt{r}}{d_{\min}} \|\mathbf{U}^\top \mathbf{E} \boldsymbol{\Theta}^\top \mathbf{V} \boldsymbol{\gamma}^*\|_\infty$$

$$\leq \frac{\sigma c_0}{d_{\min}} \|\boldsymbol{\gamma}^*\|_2 \sqrt{\frac{r \log r}{(K-1)\tau_{subs}}}.$$

**(III).** Since each entry of $\mathbf{E}$ is an independent Gaussian with variance bounded by $\sigma^2$, $\mathbb{E}\left[\mathbf{E}\mathbf{E}^\top\right] = \sigma^2(K-1)\tau_{subs}\mathbf{I}$. By Lemma 6 we have with probability $1 - c_1 \exp(-c_2 \log r)$

$$\|\mathbf{D}^{-1}\mathbf{U}^\top \left(\mathbf{E}\mathbf{E}^\top - \sigma^2(K-1)\tau_{subs}\mathbf{I}\right) \mathbf{U}\mathbf{D}^{-1}\gamma^*\|_2 \leq \frac{\sigma^2 c_0}{d_{\min}^2}\|\gamma^*\|_2 \sqrt{r \log r(K-1)\tau_{subs}}$$

**(IV).**

$$\|\mathbf{D}^{-1}\mathbf{U}^\top \left(\sigma^2(K-1)\tau_{subs}\mathbf{I}\right) \mathbf{U}\mathbf{D}^{-1}\gamma^*\|_2 \leq \frac{\sigma^2(K-1)\tau_{subs}}{d_{\min}^2}\|\gamma^*\|_2 \quad (11)$$

Combining (I) – (IV) and using $c_0 \sqrt{\frac{r \log r}{(K-1)\tau_{subs}}} \leq c' \rho_1 = \rho$ where $\rho = C\sqrt{\frac{r \log(2r/\xi)}{\tau_K}}$ and $\tau_K = (K-1)\tau_{subs}$, we have with probability $1 - \left(3c_1 \exp(-c_2 \log r) + \xi + \frac{p-\tau}{e^r}\right)$

$$(\tilde{\boldsymbol{\alpha}} - \boldsymbol{\alpha}^*)^\top (\mathbf{K} - \tilde{\mathbf{K}})\boldsymbol{\alpha}^* \leq \|\tilde{\gamma} - \gamma^*\|_2 \|\gamma^*\|_2 \rho \left(1 + 2\frac{\sigma}{d_{\min}} + \frac{\sigma^2}{d_{\min}^2}\left(\tau_K + \tau_K^2\right)\right) \quad (12)$$

On the r.h.s. of (18) we have with $a = \tilde{\gamma} - \gamma^*$

$$(\tilde{\boldsymbol{\alpha}} - \boldsymbol{\alpha}^*)^\top \tilde{\mathbf{K}}(\tilde{\boldsymbol{\alpha}} - \boldsymbol{\alpha}^*) = a^\top (\mathbf{V}^\top \Theta + \mathbf{D}^{-1}\mathbf{U}^\top \mathbf{E})(\mathbf{V}^\top \Theta + \mathbf{D}^{-1}\mathbf{U}^\top \mathbf{E})^\top a.$$

Denoting $\mathbf{w} = \Theta^\top \mathbf{V} a$ and $\mathbf{m} = \mathbf{E}^\top \mathbf{U}\mathbf{D}^{-1} a$ we have

$$(\tilde{\boldsymbol{\alpha}} - \boldsymbol{\alpha}^*)^\top \tilde{\mathbf{K}}(\tilde{\boldsymbol{\alpha}} - \boldsymbol{\alpha}^*) = (\mathbf{w} + \mathbf{m})^\top (\mathbf{w} + \mathbf{m}).$$

For convenience say $\tilde{\tau} = \tau + (K-1)\tau_{subs} = \tau + \tau_K$. Importantly, $\mathbf{m}$ is symmetric around 0, so

$$(\mathbf{w} + \mathbf{m})^\top (\mathbf{w} + \mathbf{m}) = \sum_{i=1}^{\tilde{\tau}}(w_i + m_i)^2 \geq \sum_{i=1}^{\tilde{\tau}} w_i^2 \cdot \mathbb{I}_{\{m_i > 0\}}. \quad (13)$$

The r.h.s. of this expression corresponds to randomly subsampling summands from $\mathbf{w}^\top \mathbf{w} = a^\top \mathbf{V}^\top \Theta \Theta^\top \mathbf{V} a$ where the subsampling scheme is defined by the non-zero summands stemming from the indicator function in Eq. (13). When only considering the non-zero summands, we can write the resulting matrix product as

$$(\tilde{\boldsymbol{\alpha}} - \boldsymbol{\alpha}^*)^\top \tilde{\mathbf{K}}(\tilde{\boldsymbol{\alpha}} - \boldsymbol{\alpha}^*) = (\mathbf{w} + \mathbf{m})^\top (\mathbf{w} + \mathbf{m}) \geq \sum_{i=1}^{\tilde{\tau}} w_i^2 \cdot \mathbb{I}_{\{m_i > 0\}} = a^\top \mathbf{V}^\top \tilde{\Theta}\tilde{\Theta}^\top \mathbf{V} a$$

where $\tilde{\Theta}$ contains the columns of $\Theta$ corresponding to the non-zero summands. In other words, $\tilde{\Theta}$ corresponds to a random projection matrix that projects to a lower-dimensional space than $\Theta$. Next, we can write

$$(\tilde{\boldsymbol{\alpha}} - \boldsymbol{\alpha}^*)^\top \tilde{\mathbf{K}}(\tilde{\boldsymbol{\alpha}} - \boldsymbol{\alpha}^*) \geq a^\top \mathbf{V}^\top \tilde{\Theta}\tilde{\Theta}^\top \mathbf{V} a = \|a\|_2^2 - \underbrace{a^\top \left(\mathbf{I} - \mathbf{V}^\top \tilde{\Theta}\tilde{\Theta}^\top \mathbf{V}\right) a}_{(V)}. \quad (14)$$

To lower bound the r.h.s. of this expression, we need to upper bound (V). We achieve this by first finding a lower bound on the number of non-zero summands in Eq. (13), i.e. on the number of columns of $\tilde{\Theta}$. Intuitively, the smaller the projection dimension realized by $\tilde{\Theta}$, the larger term (V) becomes. Using the Chernoff bound from Lemma 7 with $\delta = 1/2$, we can bound the probability that the number of non-zero summands lies below $\tilde{\tau}/4$ by $\exp(-\tilde{\tau}/16)$. We can then upper bound (V) using Lemma 5 for $\tilde{\Theta} \in \mathbb{R}^{p \times \tilde{\tau}/4}$. So with $\tilde{\tau} = \tau + \tau_K$ and $\tilde{\rho} = \sqrt{\frac{cr \log(2r/\xi)}{\tau/4 + (K-1)\tau_{subs}/4}}$, we have with probability $1 - (\xi + \frac{p}{e^r} + e^{-(\tau+\tau_K)/16})$

$$
\begin{aligned}
(\tilde{\boldsymbol{\alpha}} - \boldsymbol{\alpha}^*)^\top \tilde{\mathbf{K}} (\tilde{\boldsymbol{\alpha}} - \boldsymbol{\alpha}^*) &\geq \|a\|_2^2 - a^\top \left( \mathbf{I} - \mathbf{V}^\top \tilde{\Theta} \tilde{\Theta}^\top \mathbf{V} \right) a \\
&\geq \|a\|_2^2 - \tilde{\rho} \|a\|_2^2 \\
&\geq (1 - \tilde{\rho}) \|a\|_2^2 \\
&\geq (1 - 2\rho) \|a\|_2^2.
\end{aligned}
\tag{15}
$$

Plugging 12 and 15 into Lemma 4 we have with probability at least
$1 - \left(3c_1 \exp(-c_2 \log r) + 2\xi + 2\frac{p}{e^r} + e^{-(\tau+\tau_K)/16}\right)$

$$
(1 - 2\rho) \|\tilde{\gamma} - \gamma^*\|_2^2 \leq \|\tilde{\gamma} - \gamma^*\|_2 \|\gamma^*\|_2 \rho \left(1 + \frac{\sigma}{d_{\min}} \left(2 + \frac{\sigma \tau_K + \sigma \tau_K^2}{d_{\min}}\right)\right).
\tag{16}
$$

Finally, with the relationship $\boldsymbol{\beta}^* = -\frac{1}{n\lambda} \mathbf{V} \gamma^*$ and $\tilde{\boldsymbol{\beta}} = -\frac{1}{n\lambda} \mathbf{V} \tilde{\gamma}$ we have $\frac{1}{n\lambda} \|\gamma^*\|_2 = \|\boldsymbol{\beta}^*\|_2$ and $\|\tilde{\boldsymbol{\beta}} - \boldsymbol{\beta}^*\|_2 = \frac{1}{n\lambda} \|\tilde{\gamma} - \gamma^*\|_2$ due to the orthonormality of $\mathbf{V}$. Thus, we obtain the following error bound for the coefficients estimated by party $k$

$$
\|\widehat{\boldsymbol{\beta}}_k - \boldsymbol{\beta}_k^*\|_2 \leq \|\tilde{\boldsymbol{\beta}} - \boldsymbol{\beta}^*\|_2 \leq \frac{\rho}{(1 - 2\rho)} \left(1 + \frac{\sigma}{d_{\min}} \left(2 + \frac{\sigma \tau_K + \sigma \tau_K^2}{d_{\min}}\right)\right) \|\boldsymbol{\beta}^*\|_2
\tag{17}
$$

which holds with probability at least $1 - \left(3c_1 \exp(-c_2 \log r) + 2\xi + 2\frac{p}{e^r} + e^{-(\tau+\tau_K)/16}\right)$. ∎

## Appendix B. Supporting results

**Lemma 4 (Adapted from Lemma 1 (Zhang et al., 2014))** *Let $\boldsymbol{\alpha}^*$ and $\tilde{\boldsymbol{\alpha}}$ be as defined in Definition 3. We obtain*

$$
\frac{1}{\lambda} (\tilde{\boldsymbol{\alpha}} - \boldsymbol{\alpha}^*)^\top \left(\mathbf{K} - \tilde{\mathbf{K}}_k\right) \boldsymbol{\alpha}^* \geq \frac{1}{\lambda} (\tilde{\boldsymbol{\alpha}} - \boldsymbol{\alpha}^*)^\top \tilde{\mathbf{K}}_k (\tilde{\boldsymbol{\alpha}} - \boldsymbol{\alpha}^*).
\tag{18}
$$

**Proof** See (Zhang et al., 2014). ∎

**Lemma 5 (Concatenating random features (Lemma 3 from Heinze et al. (2014)))** *Consider the singular value decomposition $\mathbf{X} = \mathbf{UDV}^\top$ where $\mathbf{U} \in \mathbb{R}^{n \times r}$ and $\mathbf{V} \in \mathbb{R}^{p \times r}$ have orthonormal columns and $\mathbf{D} \in \mathbb{R}^{r \times r}$ is diagonal; $r = \text{rank}(\mathbf{X})$. In addition to the raw features, let $\bar{\mathbf{X}}_k \in \mathbb{R}^{n \times (\tau + (K-1)\tau_{subs})}$ contain random features which result from concatenating the $K - 1$ random projections from the other parties. Furthermore, assume without loss* of generality that the problem is permuted so that the raw features of party $k$'s problem are the first $\tau$ columns of $\mathbf{X}$ and $\bar{\mathbf{X}}_k$. Finally, let

$$\Theta_C = \begin{bmatrix} \mathbf{I}_\tau & 0 & \dots & 0 \\ 0 & \mathbf{\Pi}_1 & 0 & \vdots \\ \vdots & \dots & \ddots & 0 \\ 0 & \dots & \dots & \mathbf{\Pi}_{K-1} \end{bmatrix} \in \mathbb{R}^{p \times (\tau + (K-1)\tau_{subs})}$$

such that $\bar{\mathbf{X}}_k = \mathbf{X}\Theta_C$.

With probability at least $1 - \left(\xi + \frac{p-\tau}{e^r}\right)$

$$\|\mathbf{V}^\top \Theta_C \Theta_C^\top \mathbf{V} - \mathbf{V}^\top \mathbf{V}\|_2 \leq \sqrt{\frac{c \log(2r/\xi) r}{(K-1)\tau_{subs}}}.$$

**Lemma 6 (Adapted from Lemma 14 from Loh and Wainwright (2012))** *If $X \in \mathbb{R}^{n \times p_1}$ and $Y \in \mathbb{R}^{n \times p_2}$ are zero-mean sub-Gaussian matrices with parameters $(\Sigma_x, \sigma_x^2)$ and $(\Sigma_y, \sigma_y^2)$ respectively. If $n \gtrsim \log p$ then*

$$\mathbb{P}\left(\|\left(\frac{Y^\top X}{n} - \mathbb{E}\left[Y^\top X\right]\right)\|_\infty \geq c_0 \sigma_x \sigma_y \sqrt{\frac{\log p}{n}}\right) \leq c_1 \exp(-c_2 \log p).$$

**Lemma 7 (Chernoff bound for sum of independent Bernoulli trials (Goemans, 2015))** *Let $X = \sum_{i=1}^n X_i$, where $X_i = 1$ with probability $p_i$ and $X_i = 0$ with probability $1 - p_i$, and all $X_i$ are independent. Let $\mu = \mathbb{E}(X) = \sum_{i=1}^n p_i$. Then*
*(i) Upper Tail:*

$$\mathbb{P}(X \geq (1+\delta)\mu) \leq e^{-\frac{\delta^2}{2+\delta}\mu} \text{ for all } \delta > 0;$$

*(ii) Lower Tail:*

$$\mathbb{P}(X \leq (1-\delta)\mu) \leq e^{-\mu\delta^2/2} \text{ for all } 0 < \delta < 1.$$

## Appendix C. Connection between $(\epsilon, \delta, \mathcal{S})$-differential privacy and regularized least-squares estimation

In the PRIDE framework, consider the unregularized local objective function for a single party $k$ when the functions $f_i$ are the squared error. From (9) and (10) we have (omitting the subscript $k$ for ease of notation)

$$\|Y - \bar{\mathbf{X}}\mathbf{b}\|_2^2 = \|Y - (\mathbf{X}\Theta + \mathbf{E})\mathbf{b}\|_2^2.$$

Denoting $\tilde{\mathbf{X}}_l, \forall\, l \neq k$ as the concatenated random features we have

$$\|Y - (\mathbf{X}\Theta + \mathbf{E})\mathbf{b}\|_2^2 = \|Y - (\mathbf{X}_k \mathbf{b}_k + \sum_{l \neq k}(\tilde{\mathbf{X}}_l + \mathbf{W}_l)\mathbf{b}_l)\|_2^2.$$

Since all of the elements in $\mathbf{W}$ are sampled i.i.d. from an independent Gaussian with variance $\sigma^2$, let us now consider taking the expectation of this expression with respect to the randomness in $\mathbf{W}$:

$$\mathbb{E}_\mathbf{W} \|Y - (\mathbf{X}_k \mathbf{b}_k + \sum_{l \neq k}(\tilde{\mathbf{X}}_l + \mathbf{W}_l)\mathbf{b}_l)\|_2^2.$$

Due to independence, we can simply consider the univariate expectation

$$\mathbb{E}_{w \sim \mathcal{N}(0,\sigma^2)} (y - (\tilde{x}_l + w)\mathbf{b}_l)^2 = \int_{-\infty}^{\infty} \frac{1}{\sqrt{2\pi\sigma^2}} \exp\left\{-\frac{w^2}{2\pi\sigma^2}\right\} \cdot (y - (\tilde{x}_l + w)\mathbf{b}_l)^2 \, dw$$
$$= (y - \tilde{x}_l \mathbf{b}_l)^2 + \sigma^2 \mathbf{b}_l^2.$$

So in $\tau + (K-1)\tau_{subs}$ dimensions we obtain a *regularized* least squares objective where the regularization is only on the $(K-1)\tau_{subs}$ *random* features

$$\|Y - (\mathbf{X}_k \mathbf{b}_k + \sum_{l \neq k} \tilde{\mathbf{X}}_l \mathbf{b}_l)\|_2^2 + \sigma^2 \sum_{l \neq k} \mathbf{b}_l^2. \tag{19}$$

The strength of the regularization is governed by the variance of the perturbation.

## Appendix D. Additional experimental details and results

### Data tables

Table D.1: Projection dimensions

|           | $K$ | $\tau = p/K$ | $0.01\tau$ | $0.05\tau$ | $0.1\tau$ | $0.2\tau$ |
|-----------|-----|--------------|------------|------------|-----------|-----------|
| SIMULATED | 2   | 200          | 2          | 10         | 20        | 40        |
| CLIMATE   | 4   | 2,592        | 26         | 130        | 259       | 518       |
| CANCER    | 4   | 500          | 5          | 25         | 50        | 100       |

Table D.2: Largest noise standard deviation $\max_k (\sigma_k)$ when $\delta = 0.05$

| $\epsilon$ | 0.1    | 0.25  | 0.5   | 0.75  | 1     | 2     | 5    | 10   | 20   |
|------------|--------|-------|-------|-------|-------|-------|------|------|------|
| SIMULATED  | 162.47 | 66.99 | 35.10 | 24.42 | 19.05 | 10.87 | 5.67 | 3.68 | 2.48 |
| CLIMATE    | 186.59 | 76.93 | 40.30 | 28.04 | 21.88 | 12.48 | 6.51 | 4.22 | 2.84 |
| CANCER     | 239.41 | 98.71 | 51.71 | 35.98 | 28.07 | 16.02 | 8.35 | 5.42 | 3.65 |

### D.1 Simulation setting

We consider $K = 2$ blocks of features. One block of features could, for instance, contain genomic data, such as measurements of single nucleotide polymorphisms (SNPs). We shall denote the set of features contained in this block by $C$. Due to its highly personal and sensitive nature, genomic data arising from techniques like SNP genotyping is hardly ever publicly available. The other block of features could hold gene expression data. We denote this second set of features by $X$. Some of the genomic features have an effect on some of the gene expression features and both sets of features contribute to the response $Y$. This results in the structure shown in Figure D.1. Due to the dependence between $C$ and $X$ one cannot obtain accurate coefficient estimates for the effect of $X$ on $Y$ when only including $X$ into the model. In such settings, PRIDE allows to adjust for the confounding effects from $C$ on $X$ while guaranteeing $(\epsilon, \delta, \mathcal{S})$-differential privacy.

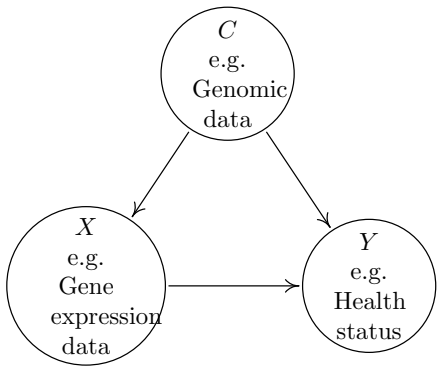

Figure D.1: The confounders in $C$ influence both the variables in $X$ as well as the response $Y$.

Specifically, each blocks of features contains $\tau = 200$ features. So $p = 400$ and we choose $n = 1000$ ($n_{train} = 800$ resp. $n_{test} = 200$). In order to create an interesting correlation structure both within the blocks of features and between $C$ and $X$, we consider a Gaussian random field on a $20 \times 20$ grid. So each grid point corresponds to one feature and we generate $n$ realizations from the model. We add confounding effects from $C$ on $X$ by selecting 20 pairs of features from $X$ and $C$ at random. Denote the set of tuples by $\mathcal{M}$ and a single tuple by $m = (i_x, j_c)$ where $i_x$ is the index of the chosen feature from $X$ and $j_c$ is the index of the chosen feature from $C$. For all tuples in $\mathcal{M}$ we set $X^{i_x} \leftarrow X^{i_x} + C^{j_c}$. Subsequently, we create the signal by aligning the coefficients $\boldsymbol{\beta}$ with the top 20 principal components of the full design matrix. Finally, the response is generated as $Y = \mathbf{X}\boldsymbol{\beta} + \boldsymbol{\eta}$. The elements of $\boldsymbol{\eta}$ are i.i.d. zero-mean Gaussian noise with a standard deviation set to achieve a signal-to-noise ratio SNR $= \|\mathbf{X}\boldsymbol{\beta}\|_2^2 / \|\boldsymbol{\eta}\|_2^2$ of approximately 0.75. In this simulation, a noise standard deviation of 500 yielded the desired SNR.

For all further details, we refer to the data generating code which is provided as a supplement to this work in the script `generate.R`.

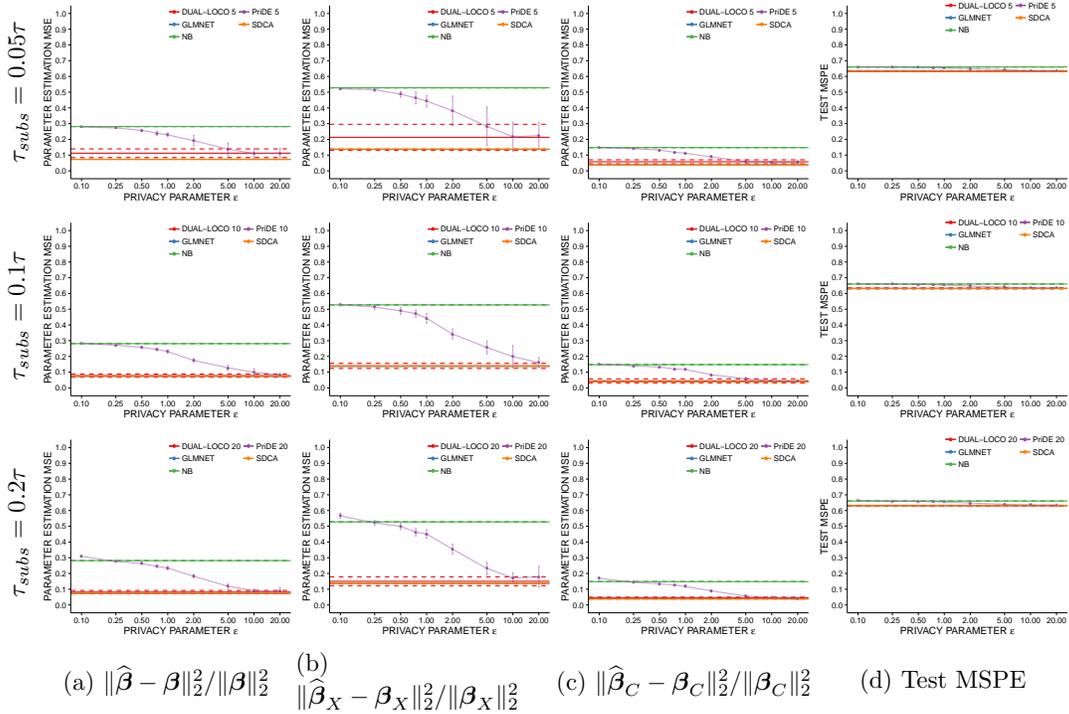

Figure D.2: **Simulated data**. Results for different projection dimensions $\tau_{subs}$. Normalized parameter estimation MSE w.r.t. true $\boldsymbol{\beta}$: (a) overall, (b) for block $X$ and (c) for block $C$. (d) Normalized prediction MSE on test set.

## D.2 Additional results for simulated data

In contrast to the other two experiments, Figure D.2 shows that the performance of PRIDE is not as sensitive to the chosen projection dimension in case of the synthetic data set. The approximation quality is fairly similar for $\tau_{subs} = \{0.05, 0.1, 0.2\} \cdot \tau$ even though the standard errors are larger for $\tau_{subs} = 0.05\tau$. This can be explained by the small value of $d_{\min}$—here, $d_{\min}$ seems to be the quantity mostly determining the term (ii) in Eq. (5), so that manipulating $\tau_{subs}$ only has a very small effect on the overall bias.

### D.3 Additional results for climate model data

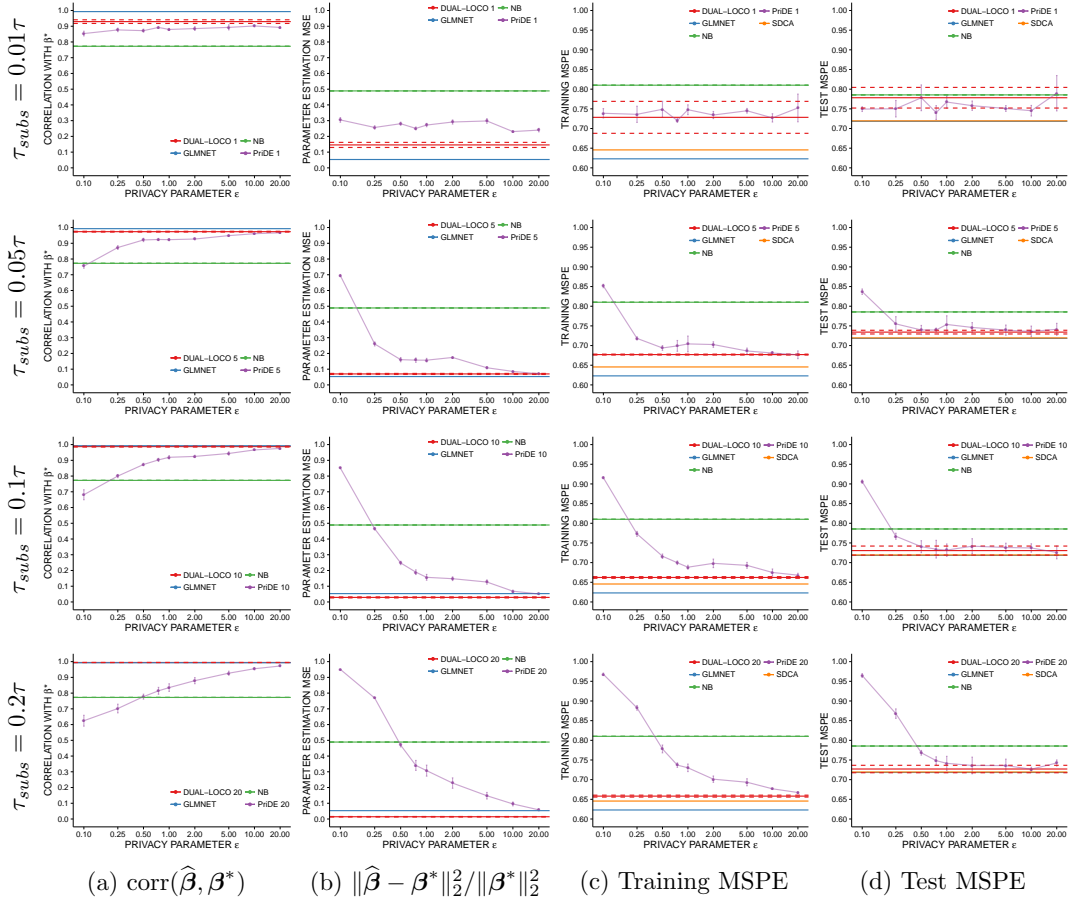

Figure D.3: **Climate model data**. Results for different projection dimensions $\tau_{subs}$. In comparison to larger projection dimensions, a projection dimension of $\tau_{subs} = 0.01\tau$ is not sufficient to capture the signal of the non-local features accurately. This is apparent from the gap in performance between DUAL-LOCO and the GLMNET/SDCA estimates which are obtained without any constraints on privacy or communication. Secondly, when $\tau_{subs} = 0.01\tau$, varying $\epsilon$ only has a very small effect on the performance of PRIDE: due to the small projection dimension the additional regularization introduced by the additive noise can be attenuated by choosing smaller values for $\lambda$ also when $\epsilon$ is small. As $\tau_{subs}$ increases, the performance of DUAL-LOCO and PRIDE improve as term (i) in Eq. (5) decreases. As predicted by Theorem 2, we also observe the adverse effect on the approximation accuracy induced by term (ii) in Eq. (5) for small values of $\epsilon$ and large values of $\tau_{subs}$.

### D.4 Additional results for gene expression data

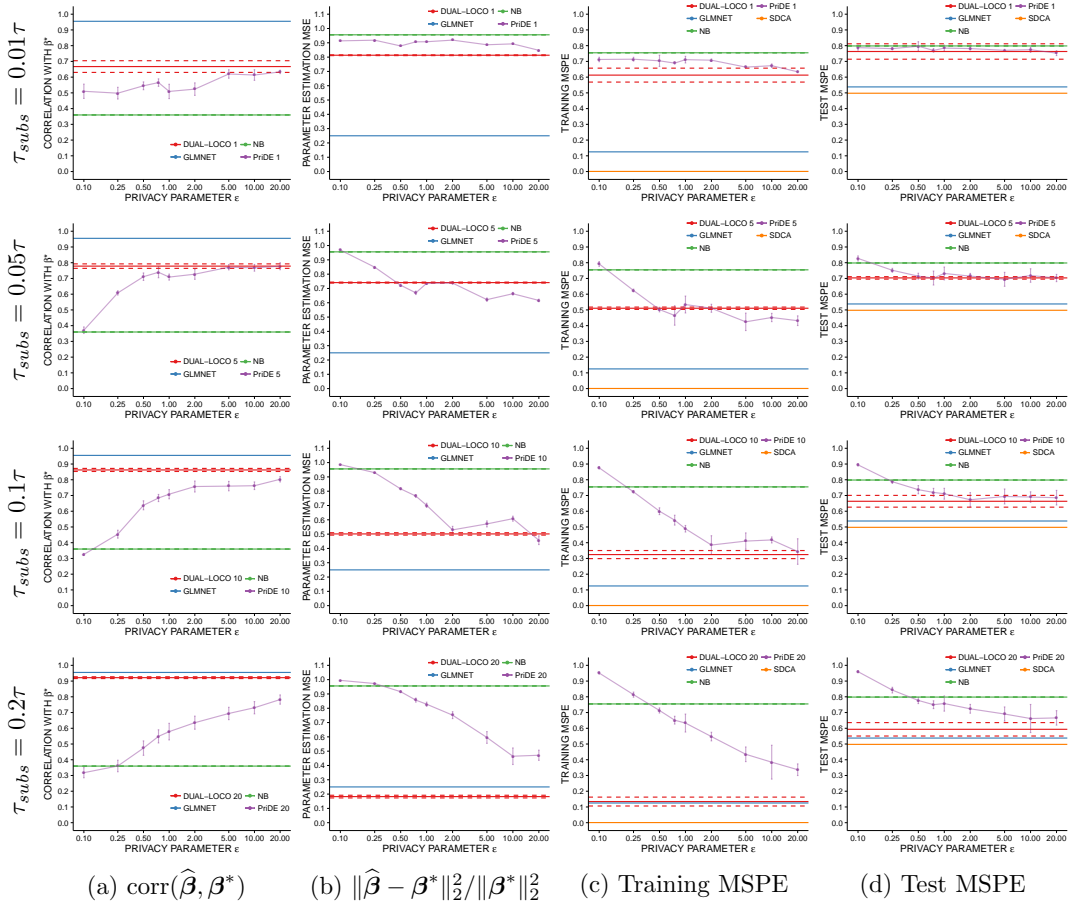

(a) corr($\widehat{\boldsymbol{\beta}}, \boldsymbol{\beta}^*$)   (b) $\|\widehat{\boldsymbol{\beta}} - \boldsymbol{\beta}^*\|_2^2 / \|\boldsymbol{\beta}^*\|_2^2$   (c) Training MSPE   (d) Test MSPE

Figure D.4: **Gene expression data**. Results for different projection dimensions $\tau_{subs}$. In comparison to larger projection dimensions, a projection dimension of $\tau_{subs} = 0.01\tau$ is not sufficient to capture the signal of the non-local features accurately. This is apparent from the gap in performance between DUAL-LOCO and the GLMNET/SDCA estimates which are obtained without any constraints on privacy or communication. Secondly, when $\tau_{subs} = 0.01\tau$, varying $\epsilon$ only has a very small effect on the performance of PRIDE: due to the small projection dimension the additional regularization introduced by the additive noise can be attenuated by choosing smaller values for $\lambda$ also when $\epsilon$ is small. As $\tau_{subs}$ increases, the performance of DUAL-LOCO and PRIDE improve as term (i) in Eq. (5) decreases. As predicted by Theorem 2, we also observe the adverse effect on the approximation accuracy induced by term (ii) in Eq. (5) for small values of $\epsilon$ and large values of $\tau_{subs}$.

## Appendix E. Privacy preserving cross validation

When the regularization parameter $\lambda$ is given, PRIDE preserves $(\epsilon, \delta, \mathcal{S})$-differential privacy (see Theorem 1). Finding a suitable $\lambda$ via $v$-fold cross validation (CV) without compromising privacy is challenging. In general, useful privacy preserving model selection procedures are an active area of research and few procedures have been proposed (Chaudhuri and Vinterbo, 2013). In Heinze et al. (2016), $\lambda$ is tuned "globally", i.e. the local predictions for a particular $\lambda$ are communicated, added and thus evaluated on the global objective. Alternatively, the local objectives could be targeted—in this case only the perturbed random features are communicated. Communicating predictions would compromise privacy so only local CV is feasible in a setting where privacy is critical. The optimal $\lambda$ is then chosen by each party individually based on the CV performance on the local design matrix, using both the raw and the perturbed random features.

A few results concerning the selection of $\lambda$ in local and global CV are given in Table E.1 which compares the chosen value for $\lambda$ using global and local cross validation on the climate dataset. For larger values of $\epsilon$, local CV selects similar values for $\lambda$ as global CV. However, for small values of $\epsilon$ ($\epsilon \leq 0.5$) the local cross validation scheme selects values for $\lambda$ that are much too large. Consequently, the predictive accuracy deteriorates, making the local CV scheme infeasible for small values of $\epsilon$. In §5, we tuned the regularization parameter using 5-fold global cross validation for all methods to assess the performance of PrIDE without confounding the comparison with this additional source of uncertainty.

One interesting aspect about the optimal value for $\lambda$ chosen by global CV is the following trend. The smaller $\epsilon$, the smaller a value for $\lambda$ tends to be selected. This is consistent with the fact that the additive noise acts as an additional regularizer (see discussion in §4.2 and §C). As this additional regularization increases with $\sigma^2$, the chosen $\lambda$ decreases, keeping the total regularization constant. However, this balancing effect is only possible as long as the additional regularization is not too large—at some point the chosen $\lambda$ approaches zero and cannot be decreased further.

Table E.1: **Cross validation results**. Comparison of the chosen value for $\lambda$ in local cross validation (LCV) and global cross validation (GCV) using the climate simulation data with projection dimension $\tau_{subs} = 0.05\tau$.

| $\epsilon$ | 0.25 | 0.5 | 0.75 | 1 | 2 | 5 | 10 | 20 |
|---|---|---|---|---|---|---|---|---|
| GCV: $\lambda$ | 47 | 125 | 137 | 140 | 159 | 116 | 108 | 104 |
| GCV: Test Mse | 0.7551 | 0.7396 | 0.7396 | 0.7537 | 0.7460 | 0.7393 | 0.7358 | 0.7414 |
| LCV: $\lambda$ for $k=1$ | $>1,000,000$ | 85,000 | 117 | 85 | 99 | 153 | 147 | 84 |
| LCV: $\lambda$ for $k=2$ | $>1,000,000$ | 90,000 | 132 | 108 | 110 | 158 | 157 | 85 |
| LCV: $\lambda$ for $k=3$ | $>1,000,000$ | 105,000 | 133 | 141 | 197 | 164 | 160 | 104 |
| LCV: $\lambda$ for $k=4$ | $>1,000,000$ | 110,000 | 174 | 145 | 237 | 250 | 161 | 111 |
| LCV: Test Mse | 0.9960 | 0.9956 | 0.7576 | 0.7399 | 0.7351 | 0.7380 | 0.7332 | 0.7329 |